%% file: main.tex
\documentclass{article}

\usepackage{microtype}
\usepackage{graphicx}
\usepackage{subfigure}
\usepackage{booktabs} 

\usepackage{hyperref}

\usepackage[accepted]{icml2024}

\usepackage{amsmath}
\usepackage{amssymb}
\usepackage{mathtools}
\usepackage{amsthm}
\usepackage{subfigure}
\usepackage[capitalize,noabbrev]{cleveref}
\usepackage{multirow}
\usepackage{threeparttable}
\usepackage{multicol,lipsum}
\usepackage{dcolumn}
\usepackage{hhline}

\theoremstyle{plain}

\theoremstyle{definition}

\theoremstyle{remark}

\usepackage[textsize=tiny]{todonotes}

\icmltitlerunning{Ranking-based Client Selection with Imitation Learning for Efficient Federated Learning}

\begin{document}

\twocolumn[
\icmltitle{Ranking-based Client Selection with Imitation Learning for Efficient Federated Learning}



\icmlsetsymbol{equal}{*}

\begin{icmlauthorlist}
\icmlauthor{Chunlin Tian}{equal,yyy}
\icmlauthor{Zhan Shi}{equal,xxx}
\icmlauthor{Xinpeng Qin}{zzz}
\icmlauthor{Li Li}{yyy}
\icmlauthor{Chengzhong Xu}{yyy}
\end{icmlauthorlist}

\icmlaffiliation{yyy}{University of Macau}
\icmlaffiliation{xxx}{University of Texas at Austin}
\icmlaffiliation{zzz}{University of Electronic Science and Technology of China}

\icmlcorrespondingauthor{Li Li}{llili@um.edu.mo}

\icmlkeywords{Machine Learning, ICML}

\vskip 0.3in
]



\printAffiliationsAndNotice{\icmlEqualContribution} 

\newcommand{\model}[1]{FedRank}

\begin{abstract}
\input{abstract}
\end{abstract}

\section{Introduction}
\input{Introduction}

\section{Background and Related Work}
\input{Background}

\section{\model~: Framework Design}
\input{Problem_Formalization}

\input{Framework}

\section{Experiments}
\input{Experiments}

\section{Conclusion}
\input{Conclusion}

\section{Impact Statements}
This paper presents work whose goal is to advance the field of Machine Learning. There are many potential societal consequences of our work, none which we feel must be specifically highlighted here.


\bibliography{main}
\bibliographystyle{icml2024}

\newpage
\appendix
\onecolumn

\input{appendix}

\end{document}

%% file: abstract.tex
Federated Learning (FL) enables multiple devices to collaboratively train a shared model while ensuring data privacy. The selection of participating devices in each training round critically affects both the model performance and training efficiency, especially given the vast heterogeneity in training capabilities and data distribution across devices. To address these challenges, we introduce a novel device selection solution called \model~, which is an end-to-end, ranking-based approach that is pre-trained by imitation learning against state-of-the-art analytical approaches. It not only considers data and system heterogeneity at runtime but also adaptively and efficiently chooses the most suitable clients for model training. Specifically, \model~ views client selection in FL as a ranking problem and employs a pairwise training strategy for the smart selection process. Additionally, an imitation learning-based approach is designed to counteract the cold-start issues often seen in state-of-the-art learning-based approaches. Experimental results reveal that \model~ boosts model accuracy by 5.2\% to 56.9\%, accelerates the training convergence up to $2.01 \times$ and save the energy consumption up to $40.1\%$.

%% file: Introduction.tex
\begin{figure*}[!ht]
    \centering
    \includegraphics[width =0.9 \linewidth]{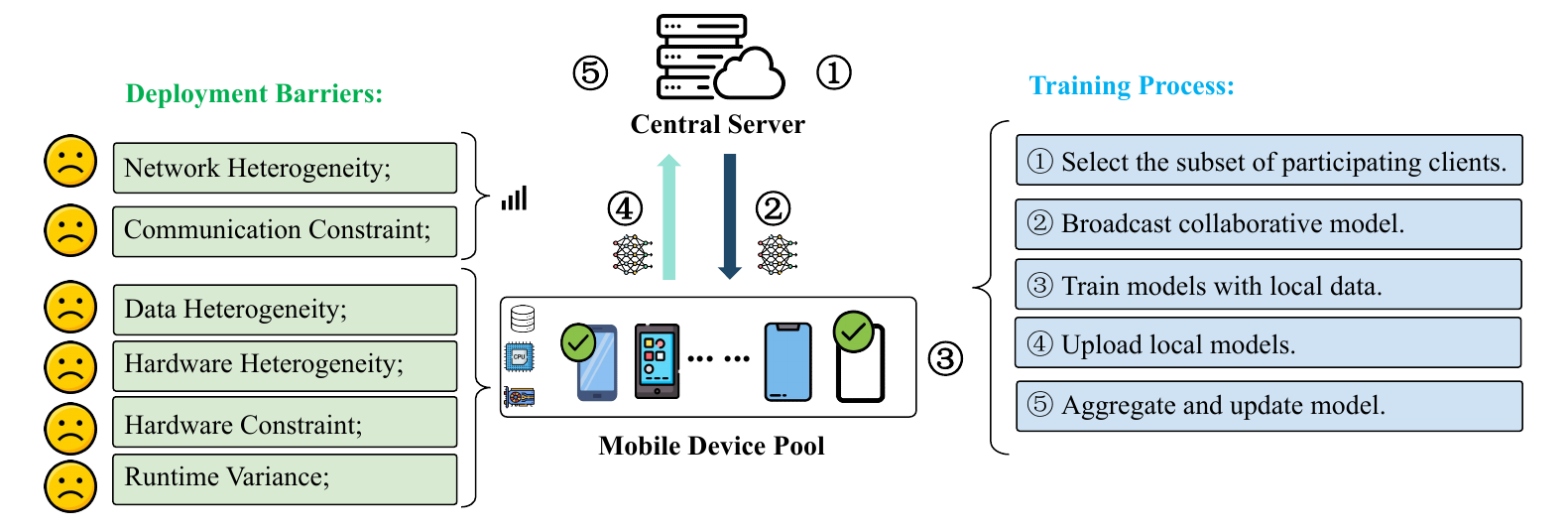}
    \caption{Workflow and real-world deployment barriers of Federated Learning.}
    \label{fig:FL_barriers}
\end{figure*}

Federated learning (FL) is a new learning paradigm that enables multiple devices to collectively train a global machine learning model while preserving data privacy~\citep{fedavg}. However, the practical deployment of FL on diverse and resource-limited mobile devices poses several challenges, as illustrated in Figure~\ref{fig:FL_barriers}. First, the disparity in training and communication capabilities of these devices, coupled with their unpredictable runtime variations, can result in stragglers. Consequently, the overall training throughput is often restricted by these low-end clients, leading to significant delays in the overall training process. Second, the data heterogeneity can influence the convergence and overall performance of the global model. In addition, computing-intensive local training and round-to-round communication with the central server can lead to substantial energy consumption, thereby affecting the battery lifespan of mobile devices.

\begin{figure}
    \centering
    \includegraphics[width =1.0\linewidth]{ 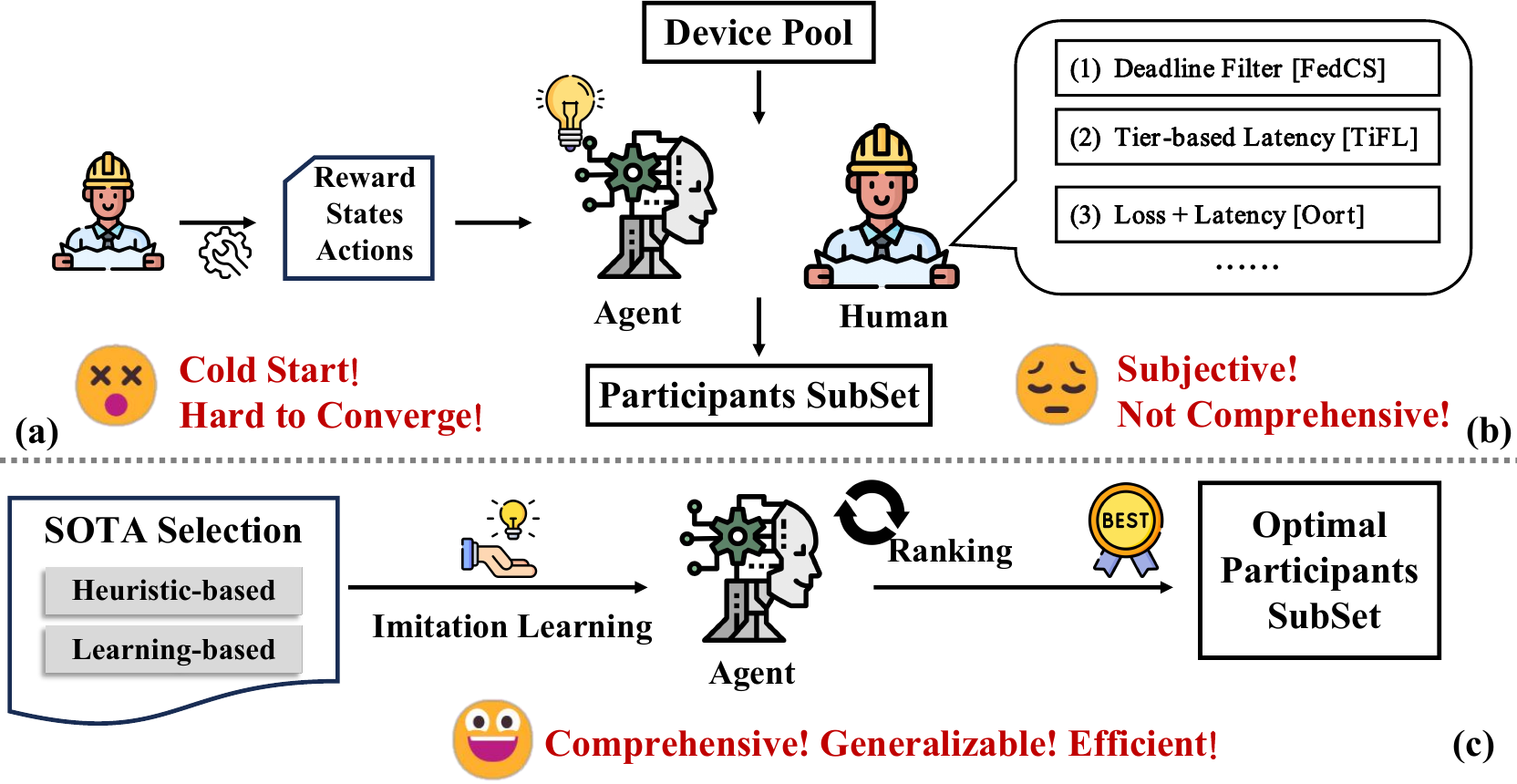}
    \caption{Illustration of \model~. (a) Learning-based approaches: tackle device selection with data-driven processes that holistically trade-off demands. (b) Heuristic-based approaches: hand-developed heuristics perform well in the specific, straightforward deployment configurations they were designed for. (c) \model~: utilizes imitation learning and a ranking approach to optimize device selection.}
    \label{fig:intro}
\end{figure}

To surmount the aforementioned barriers, the strategic selection of devices for each training round becomes pivotal to both training efficiency and model performance, as depicted in Figure~\ref{fig:intro}. 
While these heuristics excel in the specific, straightforward deployment scenarios they were designed for, they only cater to a limited subset of all potential deployment situations. As a result, their performance can be suboptimal on devices subjected to larger-scale deployments with greater diversity and complexity. Adapting to unfamiliar and intricate scenarios frequently necessitates profound domain expertise and extensive tuning.
Conversely, certain studies~\cite{RFL-1,RFL-2,multi,Tian_LTR} explore the use of reinforcement learning (RL) techniques to address device selection in a manner that holistically balances requirements using data-driven methods. Yet, real-world deployment scenarios feature a plethora of devices that constantly evolve. Such endeavors prove sample-inefficient and are plagued by the cold-start problem.

This paper presents \model~, a novel methodology for automated device selection in FL. We augment the current state-of-the-art in two pivotal ways. First, prior work typically assess each candidate device \textit{in isolation} and predict their potential value score of devices when selected. While this approach aims to accurately predict the absolute artificial score, it is not trained directly to determine which device outperforms another---the essence of the problem. In response, \textbf{we draw parallels between device selection and the learning to rank (LTR) paradigm}~\cite{LTR_3,LTR_2,LTR_1}. Our objective is to mold \model~ into a recommendation system, guiding the FL training agent toward the most impactful devices. To realize this vision, the recommendation system is devised to rank devices based on their significance, selecting the top-K. Embracing this perspective, we transition from a pointwise loss framework to a pairwise one~\cite{pairwise_1,LTR_4}, yielding pronounced advantages.
Second, we identify that the \textit{cold-start} issue has impeded the real-world utility of RL-based solutions, resulting in performance that is not as robust as that of analytical-based methodologies. To counteract this challenge, \textbf{we introduce a pre-training regimen that leverages a state-of-the-art analytical model through imitation learning (IL)}~\cite{IL_1,IL_2}. Subsequently, we refine \model~ by further training with real-world interactions, enabling it to outstrip the analytical model's performance. 
In particular, this paper makes the following contributions:
\begin{itemize}
\item This paper casts device selection in federated learning as a ranking challenge, underscoring the superiority of pairwise training compared to previous methodologies.

\item This paper proposes an offline pre-training scheme against state-of-the-art analytical solutions using imitation learning. This strategy eliminates the cold-start dilemma that has traditionally hindered the efficacy of learning-based device selection techniques.

\item Building on these strengths, our proposed method, \model~, adeptly synchronizes model performance, convergence, and energy efficiency amidst a dynamic and varied training environment. The efficacy of \model~ is substantiated through comprehensive experiments.
\end{itemize}

%% file: Background.tex
\subsection{Federated Learning} 
Figure~\ref{fig:FL_barriers} provides a visualization of a typical federated learning framework, highlighting challenges that arise during real-world implementations. At the beginning of each training cycle, a central server selects a subset $K$ of devices from the available pool $N$. It then distributes the initialized collaborative models to the chosen devices. Upon receiving the model, these selected devices conduct local training with their own data. Post-training, these devices transmit model updates back to the central server. The server then combines these updates, refining the overarching collaborative model. This cyclical process persists until the model either converges to a desired state or attains the target accuracy. Nonetheless, the variability in resources—including hardware specifications, data volume, and communication capacities—combined with stochastic runtime discrepancies can introduce inefficiencies, potentially compromising both training speed and model accuracy.

To surmount these challenges, the selection of adept training devices becomes pivotal. Prior solutions have gravitated towards two predominant strategies: heuristic-based and learning-based approaches.

\textbf{\textit{Heuristic-based selection.}} Traditional methods for device selection predominantly rely on heuristics rooted in isolated considerations such as data heterogeneity~\cite{cho2020client,heter_data,Tam_FL}, distinct training processes~\cite{train_process,tam_FL_2,ma2024,liao2024bat}, and energy efficiency~\cite{energy_efficient}. Advanced policies~\cite{oort,harmony}, have secured state-of-the-art results by employing analytical strategies that comprehensively address the multifaceted nature of device selection. Nevertheless, heuristic-based methodologies inherently risk escalating into NP-hard problems, especially with the expansive scale of FL deployments. Adapting these heuristics to unseen scenarios often demands a confluence of domain-specific expertise and extensive tuning.

\textbf{\textit{Learning-based selection.}} Several contemporary approaches have pivoted towards devising policies for device selection through learning mechanisms. Specifically, AutoFL~\citep{autofl}, Favor~\cite{RFL-2} and FedMarl~\citep{multi} employ reinforcement learning methodologies~\cite{reinforcement} to formulate decisions as Markov decision processes (MDPs). While AutoFL leverages a Q-table, Favor adopts Q-learning, and FedMarl utilizes Multi-Agent Reinforcement Learning (MARL) to achieve its objectives. In theory, these approaches are designed to select devices adeptly, however, uniformly demonstrating a tendency towards sample inefficiency are susceptible to the cold-start conundrum. This limitation often results in subpar selections during the initial phases. Consequently, the practical applicability of these methods in real-world scenarios remains circumscribed.

\subsection{Imitation Learning \& Learn to Rank}
\textit{All of above approaches (heuristic-based and learning-based) struggle in diverse and heterogeneous real-world deployment environments. Therefore how can we complement each other in order to select a qualitatively more effective device?} 

\textbf{\textit{Imitation learning.}}
As a result, we introduce imitation learning~\cite{IL_1,IL_2} to utilize well-known heuristics as conditions (expert policy) to warm-up RL (supervised). Imitation learning is instructed by the expert's demonstration, assuming its optimal settings to learn a policy, so the current policy resembles the expert one. In the IL environment, there are two peculiarities: One is that the analytical policy can be queried at any training state, which is more efficient than an expensive expert during early stages of RL. The ability to probe the expert arbitrarily allows us to avoid the performance gap with the straggler. The other is that the distribution over actions of the analytical policy is made available, enabling more sophisticated loss functions. Previous work also investigates settings with these two properties, albeit in different domains. \citet{IL_4} shows that an approximate oracle can be computed in some natural language sequence generation tasks. \citet{IL_3} learns to imitate Belady's, an oracle cache replacement policy. 

\textbf{\textit{Learn to rank.}}
For an additional dimension, traditional works typically evaluated each candidate device in isolation and predicted the potential value score of the device if it were to be selected. While this pointwise approach aims to accurately predict the absolute artificial score, it is not directly trained to determine which device is superior to another, resulting in a loss function that may overemphasize non-critical devices with low performance. Learn to Rank~\cite{LTR_2,LTR_1,LTR_3} is the application of machine learning techniques for the creation of ranking models for information retrieval systems. Ranking models typically work by predicting a relevance score $s = f(x)$ for each input $x_{i}= (q,d)$ where $q$ is a query and $d$ is a document. Once the relevance of each document is known, it can be sorted (i.e., ranked) according to these scores to find contents of interest with respect to a query. Therefore, in this paper, we introduce pairwise~\cite{pairwise_1,LTR_4} to enable the selected devices to score significantly higher than the remaining devices, which is computed as the sum of the loss terms defined on each pair of selected devices $d_i$, $d_j$, for $i, j = 1 ... n$. The goal of training the model is to predict whether the ground truth $y_i>y_j$ or not, which of two devices is more relevant.

%% file: Problem_Formalization.tex
In this section, we introduce the design of FedRank. Specifically, we first cast the problem of client selection in FL as imitation learning through framing it as learning a policy within an episodic Markov decision process. After that, we describe the integration of imitation learning as a countermeasure to the cold-start issue. Finally, we introduce the learn-to-rank methodology to effectively augment the training performance.

\subsection{Casting Device Selection as Imitation
Learning}
This serves as a foundation to outline the optimization space for \model~ that devices depend on a simple two-layer multi-layer perceptron (MLP) at the central server to perform the participation decision-making. The imitation learning phase aims to extract device selection actions $A$ based on current states $S$, which are trained with the Value Decomposition Network (VDN)~\cite{VDN} to maximize the reward $R$.

\textbf{STATE:} Specifically, the state at the $t$-th timestep $s_t = (s_t^T, s_t^E, s_t^D) \in S$ consists of three components, where: 
\begin{itemize}
\vspace{-2mm}
    \item $s_t^T = (T_{comp,i}^{t}, T_{comm,i}^{t})$ is the local training latency, consisting of training latency $T_{comp,i}^{t}$, communication latency $T_{comm,i}^{t}$, to capture system heterogeneity.
    \item $s_t^E = (E_{comp,i}^{t}, E_{comm,i}^{t})$ is the energy cost of local training, consisting of computation energy cost $E_{comp,i}^{t}$ and communication energy cost $E_{comm,i}^{t}$, to reflect system heterogeneity.
    \item $s_t^D = (L_{i}^{t}, D_{i}^{t})$ describes the data heterogeneity, where $L_i$ is the training loss of device $i$ and $D_i$ is its data size. 
\vspace{-2mm}
\end{itemize}

\textit{How to obtain device state cost-effectively is the critical challenge.} Existing approaches~\cite{cho2020client,RFL-2,oort,harmony,yebo} typically retrieve states that first force devices to complete all their local training along with updating the local DNN weights, resulting in significant processing latency. To address this issue, we introduce a “\textit{early exit}” scheme that obtains the device states after the first epoch of local training (called “probing”) processes at each device.  Probing training is performed as an intermediate result with no additional computational overhead. In addition, sending the probing training results to the central server avoids introducing extra processing latency and communication costs, since the probing states (scalars) are tiny compared to the DNN model. On the contrary, this scheme will reduce the overall processing latency and communication overhead. This is for the reason that after the outlier devices exit, only a subset of devices need to complete their local training and send their weight updates in each training round.

\textbf{ACTION:} The action set $A_t$ are $N$ binary indicators representing whether each of the $N$ devices available at timestamp $t$ should be involved in ($a_{i}^{t}=1$) or excluded from the incoming training round ($a_{i}^{t}=0$). 

\textbf{REWARD:} To characterize the primary optimization axes, we define three rewards: $R_{acc}$, $R_{T}$, and $R_{E}$ denote the test accuracy of the global model, the processing latency and the total energy consumption of each round, respectively. Collectively, these rewards strike a balance among model performance, training efficiency, and energy-friendly during the training round $t$. They are denoted as: 
\begin{equation}
    \begin{aligned}
        R^{t} &= \Delta R_{acc}^{t} \times  (\frac{T}{R_{T}^{t}})^{\textbf{1}(T < R_{T}^{t})\times \alpha} \times (\frac{E}{R_{E}^{t}})^{\textbf{1}(E < R_{E}^{t})\times \beta} \\
        &\text{where:} \\
        R_{T}^{t} &= T_{prob} + \max_{i\in K}\{[T_{comm,i}^{t}+T_{comp,i}^{t}*(l_{ep}-1)]*a_{i}^{t}\}\\
        R_{E}^{t} &= E_{prob} +\sum_{K}\{[E_{comm,i}^{t}+E_{comp,i}^{t}*(l_{ep}-1)]*a_{i}^{t}\}
        \label{opti_problem}
    \end{aligned}
\end{equation}

Where $T$ and $E$ are the developer-preferred duration and energy budget of the devices, respectively. $l_{ep}$ is the number of local training epochs. $T_{prob}$ and $E_{prob}$ represent the latency and energy cost of early exit from the first local probing epoch, respectively.
$\textbf{1}(x)$ is an indicator function that takes the value 1 if $x$ is true and 0 otherwise. In this way, the utility of those clients that may be the bottleneck of the desired speed and energy cost of the current round is penalized by a developer-specified factor $\alpha$ and $\beta$. However, we do not reward the non-straggler and non-overloader clients because their completions do not affect the effectiveness of the training round.
 
Despite the aforementioned optimization function setup employed in this paper, \model~ is highly flexible and configurable, allowing users to tailor the setup to diverse exploration objectives. Our objective is to conduct the optimal participant selection, which entails maximizing model performance in terms of accuracy and optimizing training efficiency in terms of both latency and energy-efficiency.

%% file: Framework.tex
\subsection{Offline Pre-Train with Imitation Learning}
Unlike prior RL solutions that are trained entirely on-the-fly~\citep{multi,RFL-2,RFL-1}, we pretrain the Q-Network in an offline manner using Behavioral cloning, a common approach in Imitation Learning, by imitating actions taken by an expert. Specifically, \model~ utilizes a three-layer Q-Network to approximate the Q-function for device selection, expressed as $Q_{i}^{\theta}(s,a)=\pi[R_{t}|s_{i}^{t}=s, a_{i}^{t}=a]$, where $t$ represents the current round, and the states $s_{i}^{t}$, which is obtained by profiling the $i$-th device, function as the input to the network. Once the Q-values are computed, \model~ selects the devices with top-K Q-value and the corresponding actions are set to $a_{i}^{t}=1$ while the others remain $a_{i}^{t}=0$. $\mathit{Q}(\textbf{s}_{t},\textbf{a}_{t})=\sum Q_{i}^{\theta}(s_{i}^{t},a_{i}^{t})$, where $\textbf{s}_{t}=\{s_{i}^{t}\}$ and $\textbf{a}_{t}=\{a_{i}^{t}\}$ are aggregated from all devices.

Algorithm~\ref{alg_1} summarizes the offline pre-train algorithm for \model~ policy $\pi_{\theta}$. The high-level scheme is to observe a set of states $B$ and then update the parameters $\theta$ to make the same device decision as the optimal state-of-the-art analytical policies $\pi^{*}$ as the experts for each state $s \in B$ via the loss function  $\mathit{L}_{\theta} (s, \pi^{*})$.
Specifically, we perform the analytical policies (Harmony~\cite{harmony}, Oort~\cite{oort}, FedMarl~\cite{multi}) with a given set of mobile devices and obtain episodes of state-action pairs to form an expert demonstration $B$ (lines 3-5). Given the states, we train \model~ with truncated backpropagation through time (lines 6-9). We first sample batches of states with the demonstration $B$ to initialize selection policy $\pi_{\theta}$. Then, we compute the the loss $\mathcal{L} = \sum_{i=k} \mathit{L}_{\theta} (\textbf{s}_{t},\pi_{*})$ and update the policy parameters $\theta$ based on the loss $\mathit{L}$. The pretraining encourages the learned selection policy $\pi_{\theta}(a_t |s_t)$ to make decisions that approximate the analytical policy $\pi^{*}$.

\begin{algorithm}[!ht]
\caption{\model~ Imitation Learning Algorithm.}
\begin{algorithmic}[1]
\STATE Initialize policy $\pi_{\theta}$
\FOR{step $= 0$ to $K$}
    \IF{step $\equiv 0 $}
        \STATE Collect dataset of state-action pairs to form a visited states $B = \{ \textbf{s}_t\}_{t=0}^T$ by perform the analytical policies on a given set of mobile devices.
    \ENDIF
    \STATE Sample states $\textbf{s}_{t}$ from $B$
    \STATE Warm up policy $\pi_{\theta}$ on sampled states.
    \STATE Compute loss $\mathcal{L} = \sum_{i=k} \mathit{L}_{\theta} (\textbf{s}_{t},\pi_{*})$
    \STATE Update policy parameters $\theta$ based on loss $\mathcal{L}$
\ENDFOR
\label{alg_1}
\end{algorithmic}
\end{algorithm}

\subsection{Online Learning Process}
In the deployment setting, we apply the pre-trained model to the online FL training optimization process. Meanwhile, we also adopt the target Q-network and the predict Q-network to enhance the adaptability of the post-IL network to new environments. Therefore, the Q-Network is trained recursively with minimum loss:
\begin{equation}
    L=\pi_{\textbf{s}^{t},\textbf{a},r,\textbf{s}^{t+1}}[r_{t}+\gamma*\sum_{i}Q_{i}^{\theta^{'}}(\textbf{s}_{i}^{t+1}, a)-Q_{i}^{\theta}(\textbf{s}_{i}^{t}, \textbf{a})]
    \label{loss_function}
\end{equation}
where $\theta^{'}$ denotes the target network parameters that are periodically copied from $\theta$ throughout the entire training phase.
When all devices have completed their DNN inference, the scheduler receives a global reward $r_{t}$ and proceeds to the next state $s_{i}^{t+1}$. The Profiler Cache is used to store the tuple of transitions $<s_{i}^{t},a_{i}^{t},s_{i}^{t+1},r_{t}>$ for each node $i$ to train the DNN.


\subsection{Pairwise Ranking Loss}
The ranking of the Q-values rather than the absolute values determines the selection of devices. The Q-Network trained to preserve orders among Q-values fits the problem better than the Q-Network trained to predict absolute Q-values. Thus, we adopt the pairwise loss~\cite{pairwise_1,LTR_4} defined on each pair of selected devices $d_i$, $d_j$, for $i, j = 1 ... n$. 
Specifically, in pairwise training, each pair of network outputs are mapped to a binary indicator, which is trained to minimize its distance to $y_{i,j}$ (=1 if $y_i>y_j$, 0 otherwise). We then redefine the RL rewards with the pairwise loss in Eq.(\ref{loss_function}) to prioritize the relative orders over the absolute values of rewards. We map the output Q-value (predict Q and target $\bar{Q}$) to probabilities using a logistic function as follows:
\begin{equation}
\begin{aligned}
&P_{i, j}=\sigma[Q\left(s_i, a_i\right)-Q\left(s_j, a_j\right)] \\
&\overline{P_{i, j}}=\sigma[\bar{Q}\left(s_i, a_i\right)-\bar{Q}\left(s_j, a_j\right)]
\end{aligned}
\end{equation}
Subsequently, we employ RankNet~\cite{ranknet} as a pairwise method, which uses a Binary Cross Entropy (BCE) loss to represent the pairwise loss of devices $i$ and $j$ as in Equation (\ref{pairwise}). We can then obtain the average pairwise loss: 
\begin{equation}
    \label{pairwise}
    L_{Rank} =-\sum_{i, j=1}^n\overline{P_{i, j}} \log P_{i, j}+\left(1-\overline{P_{i, j}}\right) \log \left(1-P_{i, j}\right)
\end{equation}
The new joint loss function of the RL DNN model can be defined as:
\begin{equation}
    L = \overline{L_{RL}} + \epsilon * \overline{L_{Rank}}
\end{equation}

%% file: Experiments.tex
\begin{table*}[!ht]
  \centering
  \tiny
  \caption{Model performance of different selection approaches about test accuracy, energy cost per round on average, and training speed per round on average. 
  \model~ achieves the best performance across all the datasets. $^{1}$ Set $\sigma =0.01$ for MNIST, $\sigma =0.1$ for others to emulate Non-IID. $^{2}$ Training round that target accuracy ($99\%$) is achieved.}
    \begin{tabular}{lc|ccc|ccc|ccc|ccc}
    \toprule
     \multicolumn{2}{c|}{\multirow{3}{*}{\begin{minipage}{1cm}
Dataset \\ \&Model
\end{minipage}}}  &   \multicolumn{3}{c|}{ID}    &     \multicolumn{9}{c}{OOD} \\ \cline{3-14}

& & \multicolumn{3}{c|}{MNIST-LeNet5 } & \multicolumn{3}{c|}{CIFAR10-Resnet18} & \multicolumn{3}{c|}{CINIC10-VGG16 } & \multicolumn{3}{c}{TinyImageNet-ShuffleNet }  \\ \cline{3-14}
 &     & Acc (\%) $\uparrow$& Energy  $\downarrow$& Speed  $\uparrow$ & Acc (\%) $\uparrow$& Energy  $\downarrow$& Speed  $\uparrow$ & Acc (\%) $\uparrow$& Energy  $\downarrow$& Speed  $\uparrow$ & Acc (\%) $\uparrow$& Energy  $\downarrow$& Speed $\uparrow$ \\ \midrule
\multirow{8}{*}{\rotatebox{90}{IID}} 
& FedAvg    & 98.84 & 100\% &  $1\times$   
            &84.78&100\%& $1\times$
            &64.78&100\%& $1\times$
            &34.62&100\% & $1\times$\\
&FedProx    & \underline{$99_{(35)}^{2}$}&91.7\%&  $1.03\times$
            &85.11&91.7\% & $1.01\times$
            &64.51&95.2\% & $1.05\times$ 
            &35.38&99.0\% & $1.04\times$\\
&AFL        & 98.61&75.7\%&  $1.10\times$
            &83.33&78.7\%& $1.08\times$
            &65.86&86.2\%& $1.07\times$
            &36.77&93.5\%& $1.15\times$\\
&TiFL       &98.91&71.9\%&  $1.19\times$
            &84.96&67.6\%& $1.11\times$
            &\underline{67.32}&76.3\%&  $1.35\times$
            &38.8&\underline{70.9\%} & $1.33\times$\\
&Oort       &$99_{(49)}$&55.8\% &  \underline{$1.26\times$}
            &85.69&55.8\%&  \underline{$1.40\times$}
            &67.07&69.4\% & $1.62\times$
            &\underline{40.57}&76.3\% & $1.51\times$\\
&Favor      &98.87&84.0\%&  $1.03\times$
            &85.98&81.3\%& $1.10\times$
            &65.47&97.1\%&  $1.15\times$
            &37.62&99.0\%& $1.24\times$\\
&FedMarl    &98.82 &\underline{48.1\%}&  $1.07\times$ 
            &\underline{86.37}&\underline{53.2\%}& $1.30\times$
            &66.74&\underline{68.0\%}& \underline{$1.86\times$}
            &39.48 &75.2\% & \underline{$1.67 \times$}\\ \midrule
            
&\textbf{\model~}  &$\textbf{99}_{(32)}$&\textbf{40.1\%} &  $\textbf{1.56}\times$
                   &\textbf{87.69}&\textbf{44.4\%}& $\textbf{1.67}\times$
                   &\textbf{68.58}&\textbf{52.9\%}& $\textbf{2.01}\times$
                   &\textbf{43.54}& \textbf{63.3\%}& $\textbf{1.81}\times$\\  
\bottomrule
\toprule
\multirow{8}{*}{\rotatebox{90}{Non-IID}} 
& FedAvg    &36.70&100\%& $1\times$
            &42.02&100\%& $1\times$
            &37.40&100\% & $1\times$
            &23.82&100\%& $1\times$\\
&FedProx    &67.67 &95.2\%&$1.04\times$
            &52.40&101\%& $1.04\times$
            &39.59&99.3\%& $1.05\times$
            &24.72&98.7\%& $1.03\times$\\
&AFL        &89.60&90.1\%& $1.13\times$
            &51.81&87.7\% & $1.10\times$
            &41.03&93.0\%& $1.17\times$
            &25.90&93.4\%& $1.08\times$\\
            
&TiFL       &91.21 &76.3\%& $1.21\times$
            &56.39&66.7\%& $1.20\times$ 
            &42.16&74.9\%& $1.34\times$
            &25.85&73.6\%&$1.27\times$ \\
            
&Oort       &\underline{92.47}&71.4\%& \underline{$1.36\times$}
            &53.51&61.7\%& \underline{$1.38\times$}
            &43.57&68.3\%& $1.46\times$
            &\underline{26.62}&71.5\%&$1.51\times$ \\
            
&Favor      &85.48&90.1\%& $1.08\times$
            &51.34&93.4\%& $1.16\times$
            &39.26&96.8\%& $1.19\times$
            &25.34&93.5\%& $1.20\times$\\
            
&FedMarl    &90.13&\underline{65.8\%}& $1.16\times$
            &\underline{56.73}&\underline{60.2\%}& $1.35\times$
            &\underline{44.16} &\underline{67.4\%}& \underline{$1.67\times$}
            &26.31&\underline{68.8\%}& \underline{$1.57\times$} \\ \midrule
            
&\textbf{\model~}   &\textbf{93.67}&\textbf{47.4\%}&$\textbf{1.48}\times$  
                    &\textbf{59.11}&\textbf{50.8\%}& $\textbf{1.78}\times$
                    &\textbf{47.07}&\textbf{55.2\%}& $\textbf{1.75}\times$
                    &\textbf{30.04}&\textbf{67.4\%}& $\textbf{1.83}\times$\\ 
\bottomrule
\end{tabular}%
\label{tab:acc}%
\end{table*}%

\subsection{Experimental Setup}
\textit{\textbf{Infrastructure.}} We evaluate the effectiveness of \model~ using a hybrid testbed with both simulation and off-the-shelf mobile devices which effectively emulates both data and system heterogeneity in real-world cases. Specifically, we first build a simulator following the server/client architecture based on PyTorch~\cite{pytorch}, in which different processes are created to emulate the central server and the participating devices with heterogeneous data distribution. 
Monsoon Power Monitor~\cite{monsoon} is utilized to monitor energy consumption during the training process. At the same time, the user interaction traces~\cite{livelab} are further integrated to emulate the concurrently running applications that impact the training capability at runtime.

\textit{\textbf{Models and datasets.}} The following representative models and datasets are utilized for evaluation. For the datasets, in-domain (ID) refers to the datasets consistent with imitation learning, and out-of-domain (OOD) indicates the invisible datasets. Specifically, MNIST~\cite{mnist} on LeNet5~\cite{lenet5} is adopted as the ID dataset. While for OOD datasets, ResNet18~\cite{resnet} on CIFAR10~\cite{cifar10}, VGG16~\cite{vgg16} on CINIC10~\cite{cinic}, ShuffleNet~\cite{shufflenet} on TinyImageNet~\cite{imagenet} for image classification are utilized. Furthermore, Dirichlet distribution $p_k \sim Dir_N(\sigma)$ ~\cite{dis} is utilized to simulate the non-IID data distribution across different devices. 



\textit{\textbf{Baselines.} }
We compare \model~ with three types of representative device selection approaches, including \textbf{A. Random selection:} (1) \textit{FedAvg}~\cite{fedavg} a vanilla framework for FL without any operation. (2) \textit{FedProx}~\cite{li2021model} dynamically tunes the randomly selected local device iterations utilizing training loss to keep the system robustness. \textbf{B. Heuristic-based selection:} (3) \textit{AFL}~\cite{afl} selects each round of participating devices with a probability conditioned on the current model, as well as the data on the client, to maximize efficiency. (4) \textit{TiFL}~\cite{tifl} selects devices with similar response latency for each round to reduce system heterogeneity. (5) \textit{Oort}~\cite{oort} selects the optimal participating devices based on the user-defined utility value that combines training loss and latency. \textbf{C. Learning-based selection:} (6) \textit{Favor}~\cite{RFL-2} adopts accuracy as an indicator to learn local weight selection to reduce Non-IID effects while speeding up convergence. (7) \textit{FedMarl}~\cite{multi} employs multi-agent reinforcement learning for device selection. It randomly chooses devices as RL agents each round, optimizing the final participant selection with accuracy, training latency, and communication cost. Specifically, we set the reward hyperparameters $\alpha=\beta =2$ in Eq.~\ref{opti_problem}. This is for the reason that FedRank jointly considers energy consumption and training time in each round in the reward function. We set the device pool $N=100$, and $K=10$ devices are selected to participate according to various selection policies, with $r=50$ training rounds and $l_{ep}=5$ local training epochs per round. (More details are given in Appendix~\ref{experiment}).

\begin{figure}
    \centering
     \subfigure[ToA: time to accuracy]{
    \includegraphics[width=0.48\linewidth]{ 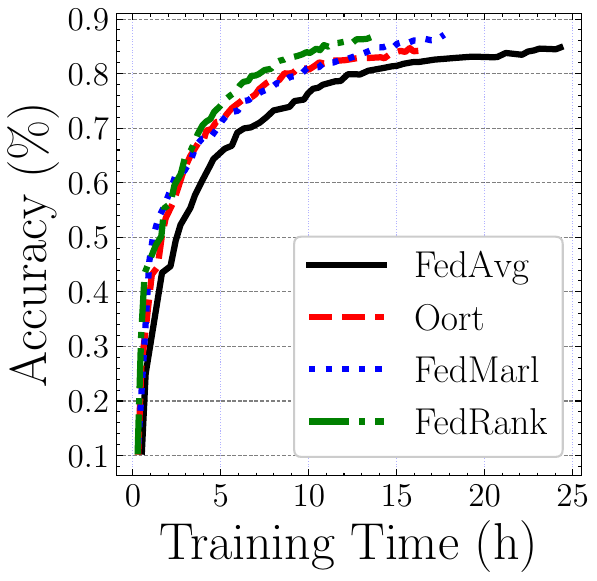}}
    \subfigure[EoA: energy to accuracy]{
    \includegraphics[width=0.48\linewidth]{ 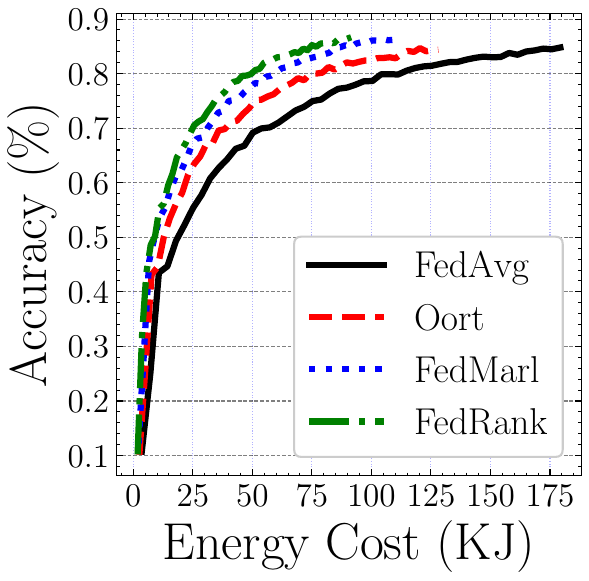}}
    \caption{Efficiency comparison of various schemes to train the ResNet18 model on CIFAR10 (IID). We compared the \model~ with the SOTA in the three baselines, respectively.}
    \label{fig:system}
\end{figure}


\subsection{Overall Performance}
\textit{\textbf{Model performance.}}
Table~\ref{tab:acc} compares the model performance of \model~ with the baselines. \model~ achieves significantly higher accuracy and converges faster than the others, 56.9\% (Non-IID) on the ID dataset. For the OOD datasets, \model~ improves the test accuracy up to 8.92\% on IID and 17.08\% on Non-IID over all datasets, while 5.2\% on IID and 11.0\% on Non-IID averaged. There are two possible reasons 1) the robust and effective groundwork laid by high-quality SOTA selection approaches through offline IL sets a solid base for the development of discerning online RL selection. 2) the pairwise loss correction term accentuates the selection or non-selection of a device, thereby elevating variance. This adjustment facilitates the model's ability to more effectively discern and assimilate the distinct attributes of devices.
Overall, this experiment demonstrates the practical and robust ability of \model~ to scale to complex workloads and application scenarios.

\textit{\textbf{System efficiency.}}
From Table~\ref{tab:acc}, we also can see that \model~ effectively speeds up the training process with faster convergence up to $2.01\times$ and achieves significant energy saving up to 40.1\%. This is for the reason that FedRank jointly considers energy consumption and training time in each round in the reward function. In contrast, no baselines are SOTA in all cases, due to incomplete consideration. Oort emphasizes solely training duration as the system effectiveness metric, while FedMarl takes into account the energy costs associated with communication. However, it overlooks the intrinsic energy overheads of the training process itself, a critical oversight, especially for devices operating on battery power.
We further conducte a comprehensive evaluation of the system efficiency of \model~ from two critical aspects: Time to Accuracy (ToA) and Energy to Accuracy (EoA). As depicted in Figure~\ref{fig:system} (a), \model~ markedly accelerates the training phase, demonstrating rapid convergence paired with enhanced accuracy. Furthermore, Figure~\ref{fig:system} (b) illustrates that \model~ also realizes considerable energy savings, underlining its efficiency. This can be attributed to \model~ thoughtfully integrates considerations of both latency and energy consumption during training. In comparison, Oort focuses exclusively on training time as the measure of system effectiveness. Conversely, while FedMarl accounts for the energy expenses related to communication, it neglects the inherent energy demands of the training phase, a significant lapse, particularly for battery-operated devices.

\begin{figure}[!ht]
    \centering
    \includegraphics[width =1.0\linewidth]{ 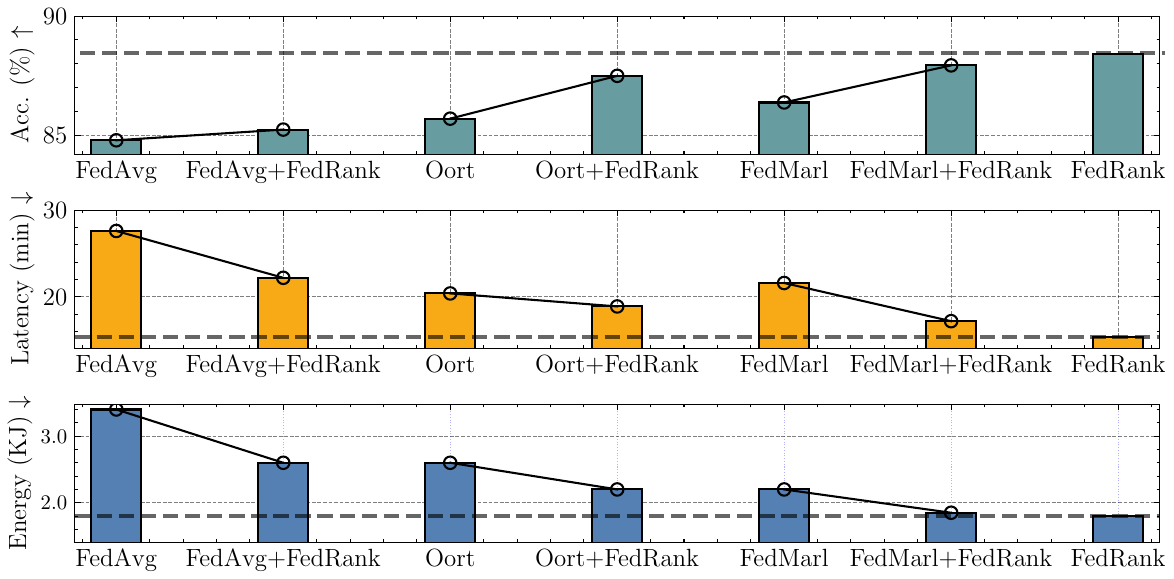}
    \vspace{-2em}
    \caption{Generalization ability of \model~ to train the ResNet18 model on CIFAR10 (IID).}
    \label{fig:gen}
\end{figure}


\textit{\textbf{Generalization and robustness.} }
An effective device selection policy must be able to suit unseen real-world deployments  (e.g., datasets and training models), as there are large amounts of mobile devices in real-world cases. Moreover, as the states are highly diverse and time-evolving, encountering all of them during imitation learning is infeasible. In this experiment, we test \model~'s ability to generalize to unseen deployments. Table~\ref{tab:acc} represents the corresponding results.  The model performance shows that with ID datasets (i.e., MNIST), \model~ improves the model convergence speed and model accuracy significantly. For OOD datasets, for the cases with simpler data and similar to domain datasets, it also shows significant performance improvement (CIFAR10). For more comprehensive datasets (CINIC10), the global training model in the baseline setting all perform inferiorly in heterogeneous real-world deployments. In contrast, \model~ maintains high accuracy even in more complicated and self-unseen scenarios, indicating that it can effectively generalize to unseen deployment environments during imitation learning.

\begin{figure}[!ht]
    \centering
    \includegraphics[width=0.8\linewidth]{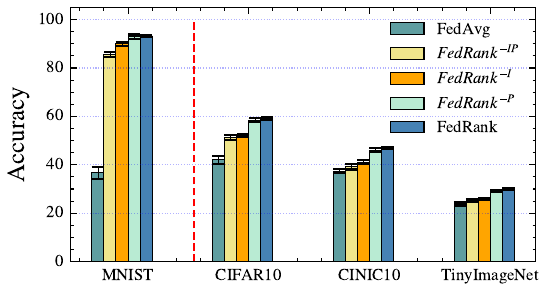}
    \caption{Ablation study explores the effects of imitation learning and pairwise loss on model performance across several datasets (MNIST, CIFAR10, CINIC, and TinyImageNet) under Non-IID settings. $\text{Model}^{-I}$ (without imitation learning), $\text{Model}^{-P}$ (without pairwise loss), and $\text{Model}^{-IP}$ (lacking both features).}
    \label{fig:perf}
\end{figure}

\subsection{Ablation Study}
To explore the impact of the different modules of \model~, we develop three model variants: (1) $\text{FedRank}^{-I}$, where offline imitation learning is ablated before FL. (2) $\text{FedRank}^{-P}$, where rank loss is ablated at DQN training. (3) $\text{FedRank}^{-IP}$, which directly employ RL to select devices.

\begin{figure}
    \centering
    \subfigure[MNIST-LeNet5, ID]{
    \includegraphics[width=0.45\linewidth]{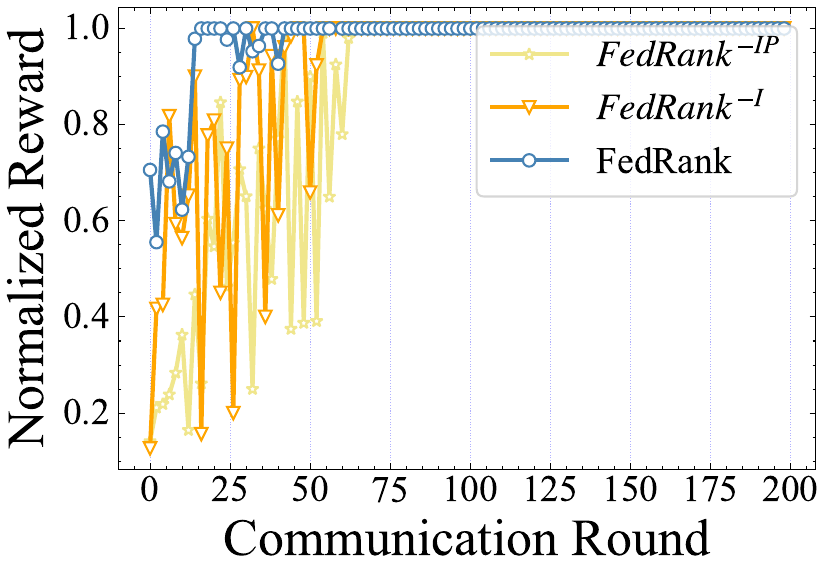}}
     \subfigure[VGG16-CIFAR10, OOD]{
    \includegraphics[width=0.45\linewidth]{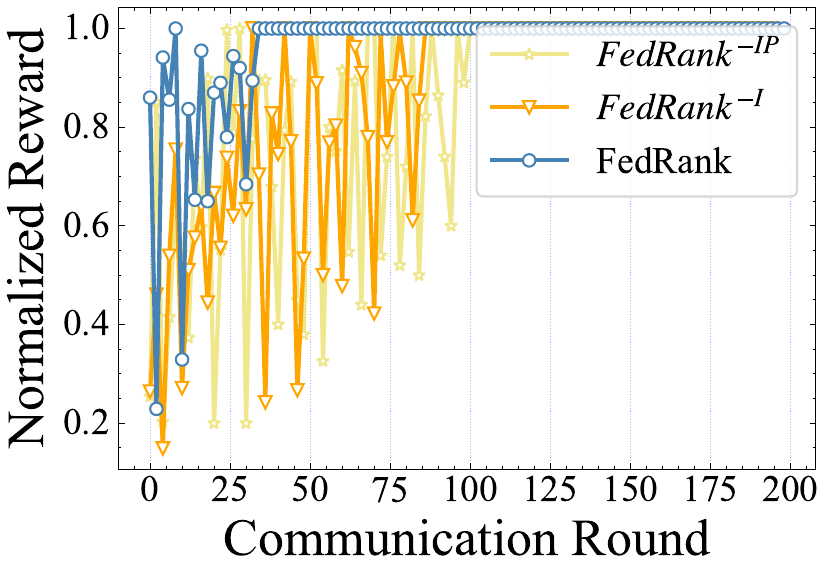}}
    \caption{Reward function analysis in the ablation study for imitation learning.}
    \label{fig:reward}
    \vspace{-1em}
\end{figure}

\textit{\textbf{Effectiveness of imitation learning.}}
Figure~\ref{fig:perf} shows that simply casting device selection as an RL problem with DQN neural architecture ($\text{\model~}^{-IP}$) only produces a negligible gain as RL faces the cold-start issue. Therefore, our additions are required to achieve better model performance with \textit{right} device selection. Meanwhile, it indicates that $\text{\model~}^{-I}$ only marginally increases efficiency in FL model training performance, while employing imitation learning can greatly enhance the accuracy by tackling the cold-start issue of the RL model in the early training round. We further present an analysis of the reward variation tendency of the RL DNN model. Figures~\ref{fig:reward} show that $\text{\model~}^{-IP}$ trained from scratch, the reward converges on average after about 70-100 aggregation rounds -- more than 200 rounds are usually required for FL convergence. Consequently, using IL to pre-train \model~ leads to faster convergence due to the rich reservoir of knowledge. 

\begin{figure} [!ht]
    \centering
    \subfigure[MNIST-LeNet5, ID]{\includegraphics[width=0.45\linewidth]{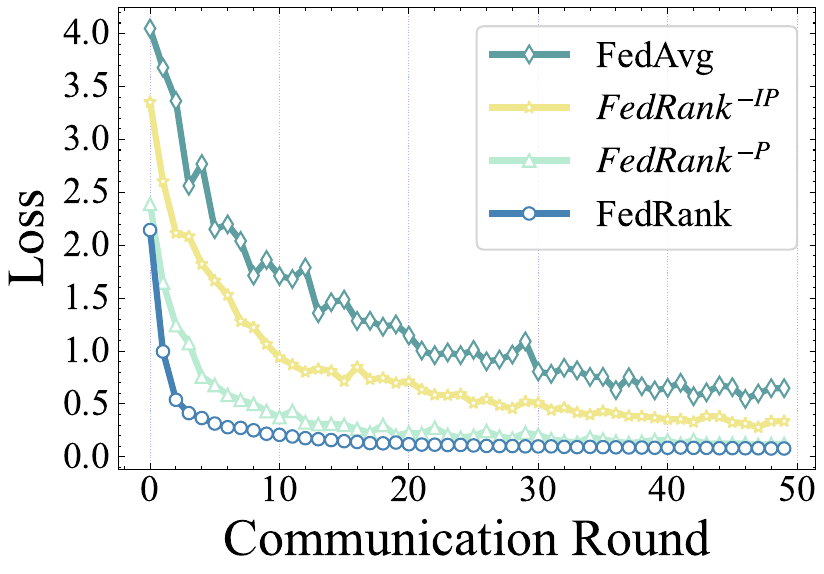}}
    \subfigure[VGG16-CIFAR10, OOD]{
    \includegraphics[width=0.45\linewidth]{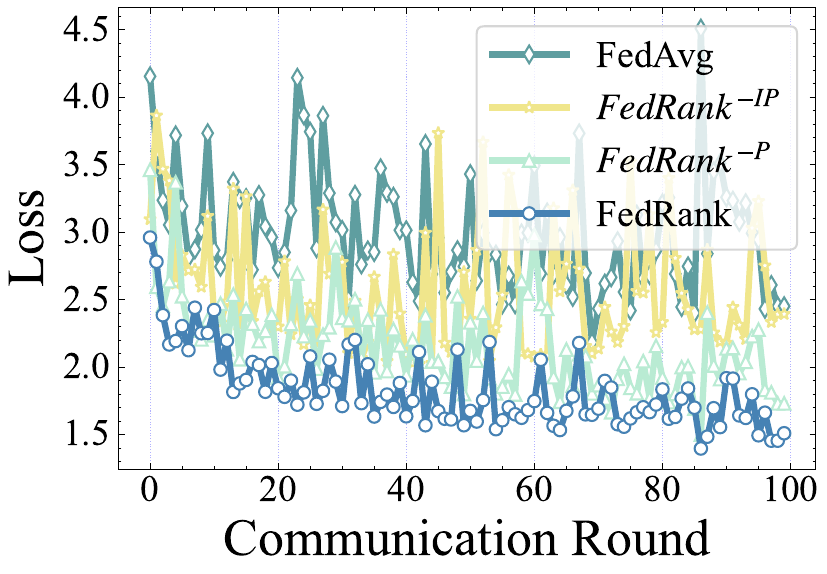}}
    \caption{Training loss analysis in the ablation study of pairwise.}
    \label{fig:loss}
\end{figure}

\begin{figure}[!ht]
     \centering 
    \includegraphics[width=0.85\linewidth]{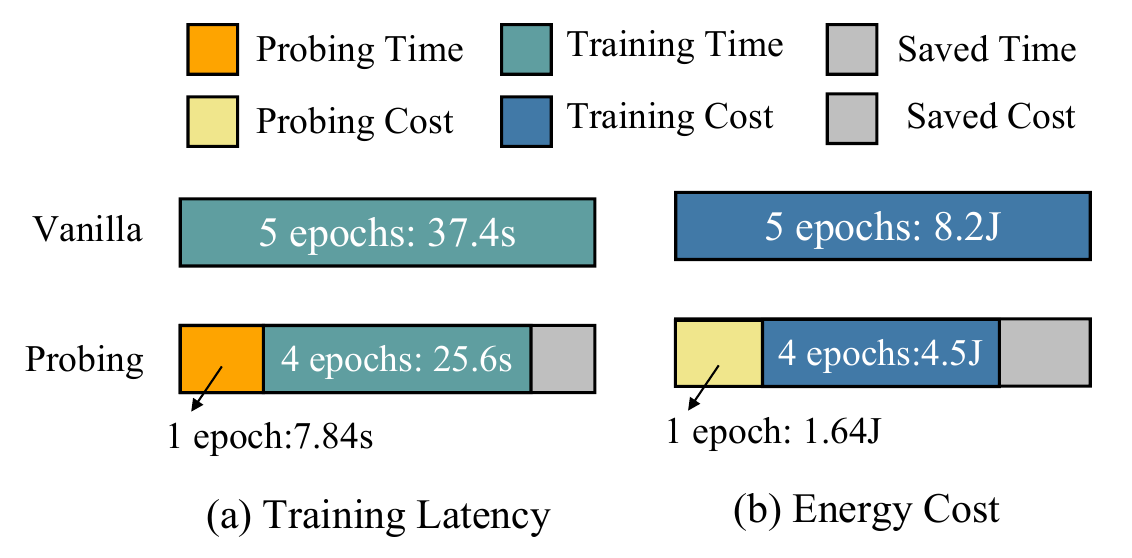}
    \caption{Average training latency and energy cost per round (5 local epochs) to train LeNet5 on MNIST. \textbf{Vanilla}: Full local training over 5 epochs to assess model bias. \textbf{Probing}: Clients perform the \textit{1st} local epoch, then report the probing loss and training overhead information to the central server. Based on this, the server performs an early rejection and stops further 4-epoch training of clients with high bias and low performance.  }
    \label{fig:probing_vs}
    \vspace{-1em}
\end{figure}

\textit{\textbf{Effectiveness of pairwise loss.} }
In addition, as Figure~\ref{fig:perf} (d) shows, the ranking loss better optimizes model performance, as it more heavily penalizes non-critical devices with lower utility values, which impairs model performance. In addition, from a distillation perspective of loss, the probability of selecting a device is proportional to the reward value for performing the ranking, which suggests that ranking loss provides greater supervision than the pointwise loss of $\text{\model~}^{-P}$. Figure~\ref{fig:loss} compares the federated learning model training convergence loss. We can observe that rank loss speeds up the training process and accelerates the model convergence rate.

\textit{\textbf{Generalization ability of \model~.}}
Figure~\ref{fig:gen} outlines when baselining the ResNet18 model via IID CIFAR10, the generalizability of \model~ in combination with different client selection methods and the effectiveness of using multiple selection methods for imitation learning.  
The adoption of a single selection method for imitation learning can significantly improve results over the corresponding baseline, highlighting the merits of continuous space paradigms in enhancing generalizability.
Furthermore, the use of diverse, multiple selection methods for imitation learning (\textit{i.e.}, the \model~) outperforms the single strategy, emphasizing the value of diversity in imitation learning for comprehensive imitation.

\begin{figure}[!ht]
    \centering
    \subfigure[$\alpha$]{\includegraphics[width=0.46\linewidth]{ 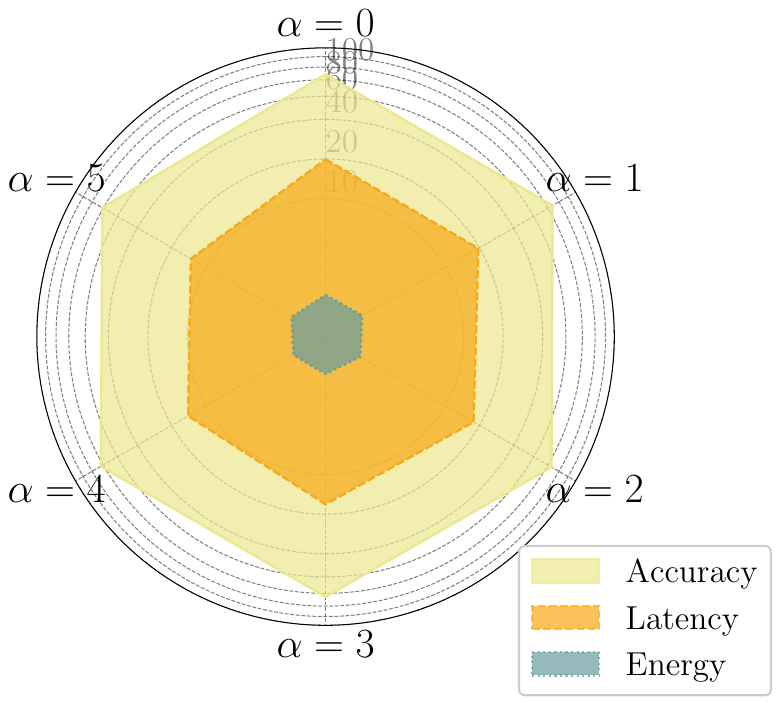}}
    \hfill
    \subfigure[$\beta$]{\includegraphics[width=0.46\linewidth]{ 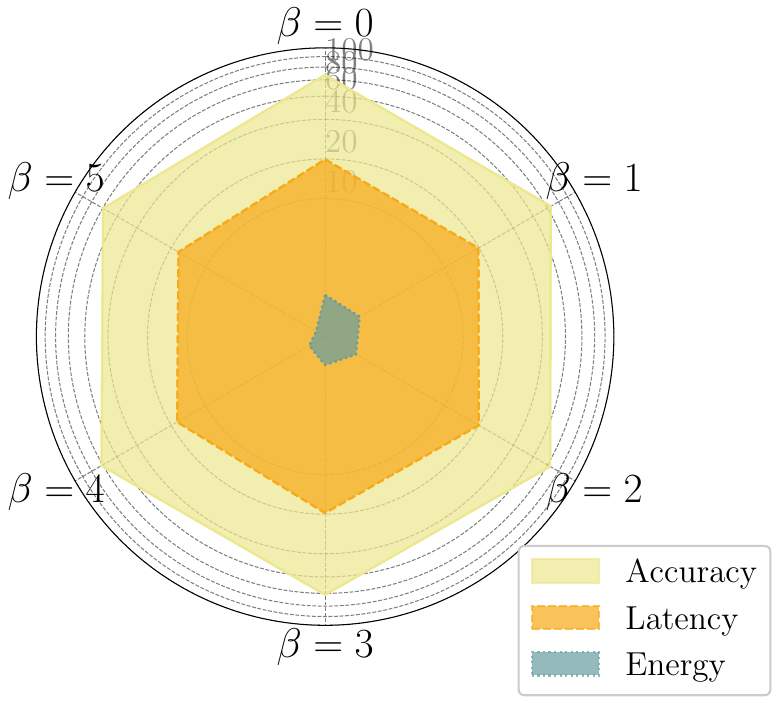}}
    \caption{Penalty factors sensitivity of \model~. $\alpha$ is the training latency penalty factor to avoid straggler, and $\beta$ is the energy penalty factor to avoid high cost devices.}
    \label{fig:sensitivity}
    \vspace{-2em}
\end{figure}

\textit{\textbf{Impact of penalty factors $\alpha$ and $\beta$ of \model~.}} \model~ uses two penalty factors to penalize the utility of high latency and energy devices in participant selection, whereby it adaptively prioritizes high system
efficiency participants. Figure~\ref{fig:sensitivity} shows that \model~ outperforms its counterparts across different $\alpha$ and $\beta$. Note that \model~ orchestrates its components to automatically navigate the best performance across parameters: larger  $\alpha$ and $\beta$ overemphasizing system efficiency drives the Pacer to relax the system constraint to admit clients with higher statistical efficiency. and vice versa. As such, \model~ achieves similar performance across all non-zero $\alpha$ and $\beta$.

\textit{\textbf{Impact of probing (early exit).}} Utilizing the Monsoon Power Monitor, we evaluate the training latency and energy consumption of the probing approach. Fig.~\ref{fig:probing_vs} shows the training latency and energy consumption for one training round under the following two schemes: 1) Vanilla and 2) Probing. Specifically, Vanilla conducts full local training over 5 epochs to access model bias. While, with probing, the clients conduct the 1st epoch of local training, then report the probing loss and training overhead to the central server.  Based on this, the server performs an early rejection and stops further local training of clients with high bias and low performance. In this case, the conventional approach consumes 37.4s of latency and 8.2J of energy in executing a local training round, effectively decreasing the latency by 10.6\% and energy consumption by 25.2\% by probing for early exit. We can see that probe training can effectively reduce resource and computational overhead and significantly improve efficiency. Meanwhile, probing data consisting only of scalars incurs negligible communication overhead compared to full model parameters.

%% file: Conclusion.tex
Federated learning is making significant strides in secure environments. Real-world FL requires selecting the right participant subset, factoring in diverse use-cases, data, systems, and dynamic changes. Our work introduces a foundational, ranking-based device selection method for efficient FL, named \model~. This method treats device selection as a ranking problem, using a pairwise training approach for intelligent selection, and continually identifying the best device group. To overcome initial challenges, \model~ employs an offline pre-training strategy, guided by advanced analytical solutions and imitation learning. The experimental results demonstrate that \model~ achieves a balance in FL between model performance and training efficiency.

%% file: appendix.tex
\section{Experiment Details}
\label{experiment}
\subsection{Dataset Details}

For four computer vision datasets, we generate IID data splits by randomly assigning training examples to each client without replacement. 
For Non-IID splits, we simulate data heterogeneity by sampling label ratios from a Dirichlet distribution $p_k \sim Dir_N(\sigma)$ with the symmetric parameter $\sigma$.
We set $\sigma =0.01$ for MNIST, and $\sigma =0.1$ for others to emulate Non-IID.
For two natural language processing datasets, Shakespeare and Wikitext are naturally non-IID.
We use LeNet5~\cite{lenet5} on MNIST, ResNet18~\cite{resnet} on CIFAR10, VGG16~\cite{vgg16} on CINIC10, ShuffleNet~\cite{shufflenet} on TinyImageNet, LSTM~\cite{LSTM} on Shakespeare. For a fair comparison, baselines and \model~ will be compared under the same settings.

\subsection{Baseline Details}
We compare \model~ with three types of representative device selection approaches, including 
\begin{itemize}
\item Random Selection:
\begin{itemize}
\item \textbf{FedAvg}~\cite{fedavg}, a vanilla framework for FL without any operation. In each round \( t \), the server randomly selects a subset of available devices \( K_t \) from the total devices \( K \). Each chosen device \( k \) trains the model on its local dataset using the current global model parameters \( w_t \), resulting in updated local parameters \( w_{t+1}^k \).
The server aggregates these updated local parameters using the formula:
\begin{equation}
    w_{t+1} = \frac{1}{K_t} \sum_{k=1}^{K_t} w_{t+1}^k
\end{equation}
This step effectively averages the local updates to form the new global model parameters.

\item \textbf{FedProx}~\cite{li2021model} improves FedAvg by handling heterogeneous data and systems, adding a proximal term to the local training objective to address non-IID data and system differences. In each round \( t \), a subset of devices \( K_t \) is selected to train on their local data with the global model parameters \( w_t \) and a proximal term. The local update \( w_{t+1}^k \) is obtained by minimizing \( L_k(w) + \frac{\mu}{2} \|w - w_t\|^2 \), where \( L_k(w) \) is the local loss, and \( \mu \) adjusts the proximal term's influence. FedProx's main advantage is its ability to stabilize training across diverse data and system environments by ensuring local updates remain close to the global model.
\end{itemize}

\item Heuristic-based Selection:
\begin{itemize}
\item \textbf{AFL}~\cite{afl} uses a device selection method conditioned on the model and client data to enhance efficiency in each round \( t \). This method focuses on devices likely to offer significant model improvements based on data diversity, model uncertainty, or past updates' impact. Post-training, devices submit their updates and informativeness measures (like loss or uncertainty) to the server. The server then combines these updates, potentially weighted by informativeness, to revise the global model to \( w_{t+1} \), the following:
\begin{equation}
   w_{t+1} = \sum_{k=1}^{K_t} \alpha_k w_{t+1}^k
\end{equation}
Here, \( \alpha_k \) denotes the weight of the \( k \)-th device's contribution.

\item \textbf{TiFL}~\cite{tifl} groups devices by response latency to mitigate system heterogeneity, sorting them into tiers by capability, bandwidth, and data quality. Each tier has a Tier Server (TS) where selected devices contribute to local updates. These updates are first aggregated within tiers at the TS using:
\begin{equation}
   w_{t+1}^{TS} = \frac{1}{K_t} \sum_{k=1}^{K_t} w_{t+1}^k
\end{equation}
Then, Tier Servers forward their updates to a central server for global aggregation, potentially weighted by tier attributes:
\begin{equation}
   w_{t+1} = \sum_{i=1}^{T} \beta_i w_{t+1}^{TS_i}
\end{equation}
This balances individual contributions and overall data characteristics.

\item \textbf{Oort}~\cite{oort} optimizes device selection by integrating training loss and latency into a user-defined utility. To enhance efficiency, it's crucial to maximize the per-unit-time statistical utility. The utility of client $i$ is defined as a product of her statistical utility and the global system utility, considering the duration of each training round, as shown in the equation:
\begin{equation}
\operatorname{Util}(i)=\underbrace{\left|B_i\right| \sqrt{\frac{1}{\left|B_i\right|} \sum_{k \in B_i} \operatorname{Loss}(k)^2}}_{\text {Statistical utility } U(i)} \times \underbrace{\left(\frac{T}{t_i}\right)^{\textbf{1}\left(T<t_i\right) \times \alpha}}_{\text {Global sys utility }}
\end{equation}
\end{itemize}

\item Learning-based Selection:
\begin{itemize}
\item \textbf{Favor}~\cite{RFL-2} utilizes accuracy to determine local weight selection, mitigating non-IID impacts and enhancing convergence. It involves training a DRL agent via a double-deep Q-learning Network (DQN). Despite evident disparities in local model weights, which hold guiding data for device selection, the DQN agent's training aims to optimize the expected total discounted reward, represented as \( R=\sum_{t=1}^T \gamma^{t-1} r_t=\sum_{t=1}^T \gamma^{t-1}\left(\Xi^{\left(\omega_t-\Omega\right)}-1\right) \).

\item \textbf{FedMarl}~\cite{multi} applies multi-agent reinforcement learning to select devices for federated learning. It strategically selects RL agents each round to optimize accuracy, training latency, and communication cost. The reward $r_t$ for each round $t$ is given by:
\begin{equation}
r_t=w_1[U(A c c(t))-U(A c c(t-1))]-w_2 H_t-w_3 B_t
\end{equation}
where $H_t$, the processing latency, is:
\begin{equation}
H_t=\max _{1 \leq n \leq N}\left(H_{t, n}^p\right)+\max _{n: 1 \leq n \leq N, a_n^t=1}\left(H_{t, n}^{r e s t}+H_{t, n}^u\right)
\end{equation}
In this, $\max _{1 \leq n \leq N} H_{t, n}^p$ signifies the maximum time for generating probing losses, used by MARL agents for client selection and model update. The time for client $n$ to complete training and upload updates is $\max _{n: 1 \leq n \leq N, a_n^t=1}\left(H_{t, n}^{\text {rest }}+\right.$ $\left. H_{t, n}^u\right)$ Here, $U(.)$ is a utility function ensuring even modest improvements in $Acc(t)$ are recognized towards the end of the learning process, and $B_t$ represents the total communication cost.

\end{itemize}
\end{itemize}


\section{Convergence Analysis}
\label{Convergence}
In this section we present a convergence analysis as a proof of the stability of \model~, based on the works in \cite{Convergence_1,multi}. 
Given that each client $n \in N$ contains $D_n$ local training data with underlying probability $P_n($.$)$. Denote $\boldsymbol{X} \times \boldsymbol{Y}$ as the compact space and label space $[C]$, where class $[C] =\{1, ..., C\}$. While $\{ \boldsymbol{x}, y_{\boldsymbol{x}}\}$ as the training data point and its ground truth label. Also, denote $f_m(\boldsymbol{x}, \boldsymbol{w})$ as the output probability for input $\boldsymbol{x}$ to take label $m$ using DNN model $f$, where $\boldsymbol{w}$ is the weight of the neural network. 
On each device, local conventional SGD is conducted separately. At timestamp $t$ on device $n \in N$, local performs: $\boldsymbol{w}_{n\_loc}^{t} =\boldsymbol{w}_{n\_loc}^{t-1}-\eta \nabla_{\boldsymbol{w}} \ell\left(\boldsymbol{w}_{n\_loc}^{t-1}\right)$.
where $\ell (\boldsymbol{w}) = \sum_{i=1}^C p^{(n)}(y=i) \mathbf{E}_{\boldsymbol{x} \mid y=i}\left[\log f_i\left(\boldsymbol{x}, \boldsymbol{w}^{t-1}_{n\_loc}\right)\right]$.
Then, let $\boldsymbol{w}_{glb}^{t}=\sum_{n=1}^N \frac{D_{n}}{\sum_{n=1}^N D_{n}} \boldsymbol{w}_{n\_loc}^{t}$ denotes the trained DNN weights after $t$-th update by aggregating the local weight. Theorem 1 provides a bound on $\left|\boldsymbol{w}_{glb}^{t}-\boldsymbol{w}_{n\_loc}^{t}\right|$.

\textbf{Theorem 1.} 
If $\nabla_{\boldsymbol{w}} E_{\boldsymbol{x} \mid y_{\boldsymbol{x}}=m} \log\left(f_m(\boldsymbol{x}, \boldsymbol{w})\right)$ is $\lambda_{\boldsymbol{x} \mid y_{\boldsymbol{x}}=m}$-Lipschitz for each $m\in M$. Assume $a_n^t$ satisfies $P(m)=\sum_{n=1}^N \frac{D_n a_n^t P_n(m)}{\sum_{n=1}^N D_n a_n^t}+\epsilon_t$ for a constant $\epsilon_t$ and $\epsilon_t \leq \epsilon$ $\forall t$. Then, we have the following inequality for the weight divergence after the $t$-th aggregation.

\begin{equation}
    \begin{aligned}
    &\left\|\boldsymbol{w}_{glb}^{t} -\boldsymbol{w}_{n\_loc}^{t}\right\|  \leq  q_n^t*[\sum_{n=1}^N a^{t}_{n}*\left\|\boldsymbol{w}^{t-1}_{glb}-\boldsymbol{w}^{t-1}_{n\_loc}\right\| \\
    & +\eta   \sum_{n=1}^N\sum_{m=1}^M\left\|P_{n}(m) - P(m)\right\| \sum_{j=1}^{t-1}a^{i}_{n}*g_{\max }\left(\boldsymbol{w}^{t-1-n}_{n\_loc}\right) ]
\end{aligned}
\end{equation}

 where $g_{\max }(\boldsymbol{w})=\max _{i=1}^C\left\|\nabla_{\boldsymbol{w}} \mathbf{E}_{\boldsymbol{x} \mid y=i} \log f_i(\boldsymbol{x}, \boldsymbol{w})\right\|$,
 $a_{n}=1+\eta \sum_{i=1}^C P_{n}(y=i) \lambda_{\boldsymbol{x} \mid y=i}$, $q_n^t = \frac{D_{n}}{\sum_{n=1}^N D_{n}}$ and $\eta$ is the learning rate. Theorem 1 states that, given certain conditions and assumptions, the local DNN model $\boldsymbol{w}_{n_loc}^{t}$ and global model $\boldsymbol{w}_{glb}^{t}$ are both bounded $\forall t \in T$. This bound implies that the stability of \model~ is maintained,  ensuring that the learning process converges. 
\subsection{Proof of Theorem 1}
Based on the definition of $\boldsymbol{w}_{glb}^{t}$ and $\boldsymbol{w}_{n\_loc}^{t}$, we have

\begin{multicols}{2}
\begin{equation*}
\centering
\begin{aligned}
& \left\|\boldsymbol{w}_{glb}^{t}-\boldsymbol{w}_{n\_loc}^{t}\right\| \\
& =\left\|\sum_{n=1}^N \frac{D_{n}}{\sum_{n=1}^N D_{n}} \boldsymbol{w}_{glb}^{t}-\boldsymbol{w}_{n\_loc}^{t}\right\| \\
& =\| \sum_{n=1}^N \frac{D_{n}}{\sum_{n=1}^N D_{n}}\left(\boldsymbol{w}_{glb}^{t}-\eta \sum_{i=1}^C p^{(k)}(y=i) \nabla_{\boldsymbol{w}} \textbf{E}_{\boldsymbol{x} \mid y=i}\left[\log f_i\left(\boldsymbol{x}, \boldsymbol{w}_{glb}^{t-1}\right)\right]\right. \\
& \left.-\boldsymbol{w}_{n\_loc}^{t-1}+\eta \sum_{i=1}^C p(y=i) \nabla_{\boldsymbol{w}} \textbf{E}_{\boldsymbol{x} \mid y=i}\left[\log f_i\left(\boldsymbol{x}, \boldsymbol{w}_{n\_loc}^{t-1}\right)\right]\right) \\
& \stackrel{1}{\leq}\left\|\sum_{n=1}^N \frac{D_{n}}{\sum_{n=1}^N D_{n}} \boldsymbol{w}_{glb}^{t-1}-\boldsymbol{w}_{n\_loc}^{t-1}\right\| \\
& +\eta\left\|\sum_{n=1}^N \frac{D_{n}}{\sum_{n=1}^N D_{n}} \sum_{i=1}^C p^{(k)}(y=i)\left(\nabla_{\boldsymbol{w}} \textbf{E}_{\boldsymbol{x} \mid y=i}\left[\log f_i\left(\boldsymbol{x}, \boldsymbol{w}_{glb}^{t-1}\right)\right]-\nabla_{\boldsymbol{w}} \textbf{E}_{\boldsymbol{x} \mid y=i}\left[\log f_i\left(\boldsymbol{x}, \boldsymbol{w}_{n\_loc}^{t-1}\right)\right]\right)\right\| \\
& \stackrel{2}{\leq} \sum_{n=1}^N \frac{D_{n}}{\sum_{n=1}^N D_{n}}\left(1+\eta \sum_{i=1}^C p^{(k)}(y=i) \lambda_{\boldsymbol{x} \mid y=i}\right)\left\|\boldsymbol{w}_{glb}^{t-1}-\boldsymbol{w}_{n\_loc}^{t-1}\right\|. \\
&
\end{aligned}
\end{equation*}
\end{multicols}

Here, inequality 1 holds because for each class $i \in[C], p(y=i)=\sum_{k=1}^K \frac{D_{n}}{\sum_{n=1}^N D_{n}} p^{(k)}(y=i)$, i.e., the data distribution over all the clients is the same as the distribution over the whole population. Inequality 2 holds because we assume $\nabla_{\boldsymbol{w}} \textbf{E}_{\boldsymbol{x} \mid y=i}\left[\log f_i(\boldsymbol{x}, \boldsymbol{w})\right]$ is $\lambda_{\boldsymbol{x} \mid y=i}$-Lipschitz.
In terms of $\left\|\boldsymbol{w}_{glb}^{t-1}-\boldsymbol{w}_{n\_loc}^{t-1}\right\|$ for client $n \in[N]$, we have

\begin{multicols}{2}
\begin{equation*}
\centering
\begin{aligned}
& \left\|\boldsymbol{w}_{glb}^{t-1}-\boldsymbol{w}_{n\_loc}^{t-1}\right\| \\
& =\| \boldsymbol{w}_{glb}^{t-2}-\eta \sum_{i=1}^C p^{(k)}(y=i) \nabla_{\boldsymbol{w}} \textbf{E}_{\boldsymbol{x} \mid y=i}\left[\log f_i\left(\boldsymbol{x}, \boldsymbol{w}_{glb}^{t-2}\right)\right] \\
& -\boldsymbol{w}_{n\_loc}^{t-2}+\eta \sum_{i=1}^C p(y=i) \nabla_{\boldsymbol{w}} \textbf{E}_{\boldsymbol{x} \mid y=i}\left[\log f_i\left(\boldsymbol{x}, \boldsymbol{w}_{glb}^{t-2}\right)\right] \mid \\
& \leq\left\|\boldsymbol{w}_{glb}^{t-2}-\boldsymbol{w}_{n\_loc}^{t-2}\right\|+\eta \| \sum_{i=1}^C p^{(k)}(y=i) \nabla_{\boldsymbol{w}} \textbf{E}_{\boldsymbol{x} \mid y=i}\left[\log f_i\left(\boldsymbol{x}, \boldsymbol{w}_{glb}^{t-2}\right)\right] \\
& -\sum_{i=1}^C p(y=i) \nabla_{\boldsymbol{w}} \textbf{E}_{\boldsymbol{x} \mid y=i}\left[\log f_i\left(\boldsymbol{x}, \boldsymbol{w}_{n\_loc}^{t-2}\right)\right] \| \\
& \stackrel{3}{\leq}\left\|\boldsymbol{w}_{glb}^{t-2}-\boldsymbol{w}_{n\_loc}^{t-2}\right\|+\eta\left\|\sum_{i=1}^C p^{(k)}(y=i)\left(\nabla_{\boldsymbol{w}} \textbf{E}_{\boldsymbol{x} \mid y=i}\left[\log f_i\left(\boldsymbol{x}, \boldsymbol{w}_{glb}^{t-2}\right)\right]-\nabla_{\boldsymbol{w}} \textbf{E}_{\boldsymbol{x} \mid y=i}\left[\log f_i\left(\boldsymbol{x}, \boldsymbol{w}_{n\_loc}^{t-2}\right)\right]\right)\right\| \\
& +\eta \| \sum_{i=1}^C\left(p^{(k)}(y=i)-p(y=i)\right) \nabla_{\boldsymbol{w}} \textbf{E}_{\boldsymbol{x} \mid y=i}\left[\log f_i\left(\boldsymbol{x}, \boldsymbol{w}_{n\_loc}^{t-2}\right)\right]|| \\
& \stackrel{4}{\leq}\left(1+\eta \sum_{i=1}^C p^{(k)}(y=i) L_{\boldsymbol{x} \mid y=i}\right)\left\|\boldsymbol{w}_{glb}^{t-2}-\boldsymbol{w}_{n\_loc}^{t-2}\right\|+\eta g_{\max }\left(\boldsymbol{w}_{n\_loc}^{t-2}\right) \sum_{i=1}^C\left\|p^{(k)}(y=i)-p(y=i)\right\| . \\
&
\end{aligned}
\end{equation*}
\end{multicols}

Here, inequality 3 holds because

\begin{multicols}{2}
\begin{equation*}
    \centering
    \begin{aligned}
& \| \sum_{i=1}^C p^{(k)}(y=i) \nabla_{\boldsymbol{w}} 	\textbf{E}_{\boldsymbol{x} \mid y=i}\left[\log f_i\left(\boldsymbol{x}, \boldsymbol{w}_{glb}^{t-2}\right)\right]-\sum_{i=1}^C p(y=i) \nabla_{\boldsymbol{w}} 	\textbf{E}_{\boldsymbol{x} \mid y=i}\left[\log f_i\left(\boldsymbol{x}, \boldsymbol{w}_{n\_loc}^{t-2}\right)\right] \mid \\
& =|| \sum_{i=1}^C p^{(k)}(y=i) \nabla_{\boldsymbol{w}} 	\textbf{E}_{\boldsymbol{x} \mid y=i}\left[\log f_i\left(\boldsymbol{x}, \boldsymbol{w}_{glb}^{t-2}\right)\right]-\sum_{i=1}^C p^{(k)}(y=i) \nabla_{\boldsymbol{w}} 	\textbf{E}_{\boldsymbol{x} \mid y=i}\left[\log f_i\left(\boldsymbol{x}, \boldsymbol{w}_{n\_loc}^{t-2}\right)\right]+ \\
& \sum_{i=1}^C p^{(k)}(y=i) \nabla_{\boldsymbol{w}} 	\textbf{E}_{\boldsymbol{x} \mid y=i}\left[\log f_i\left(\boldsymbol{x}, \boldsymbol{w}_{n\_loc}^{t-2}\right)\right]-\sum_{i=1}^C p(y=i) \nabla_{\boldsymbol{w}} 	\textbf{E}_{\boldsymbol{x} \mid y=i}\left[\log f_i\left(\boldsymbol{x}, \boldsymbol{w}_{n\_loc}^{t-2}\right)\right]|| \\
& \leq|| \sum_{i=1}^C p^{(k)}(y=i)\left(\nabla_{\boldsymbol{w}} 	\textbf{E}_{\boldsymbol{x} \mid y=i}\left[\log f_i\left(\boldsymbol{x}, \boldsymbol{w}_{glb}^{t-2}\right)\right]-\nabla_{\boldsymbol{w}} 	\textbf{E}_{\boldsymbol{x} \mid y=i}\left[\log f_i\left(\boldsymbol{x}, \boldsymbol{w}_{n\_loc}^{t-2}\right)\right]\right) \| \\
& +\left\|\sum_{i=1}^C\left(p^{(k)}(y=i)-p(y=i)\right) \nabla_{\boldsymbol{w}} 	\textbf{E}_{\boldsymbol{x} \mid y=i}\left[\log f_i\left(\boldsymbol{x}, \boldsymbol{w}_{n\_loc}^{t-2}\right)\right]\right\| \\
&
\end{aligned}
\end{equation*}
\end{multicols}

Inequality 4 holds because $g_{\max }\left(\boldsymbol{w}_{glb}^{t-2}\right)=\max _{i=1}^C\left\|\nabla_{\boldsymbol{w}} \textbf{E}_{\boldsymbol{x} \mid y=i} \log f_i\left(\boldsymbol{x}, \boldsymbol{w}_{glb}^{t-2}\right)\right\|$.
Based on Eq. (3), let $a_{n}=1+\eta \sum_{i=1}^C p^{(k)}(y=i) \lambda_{\boldsymbol{x} \mid y=i}$, by induction, we have

\begin{multicols}{2}
\begin{equation*}
    \centering
\begin{aligned}
& \left\|\boldsymbol{w}_{glb}^{t-1}-\boldsymbol{w}_{m T-1}^{(c)}\right\| \\
\leq & a_{n}\left\|\boldsymbol{w}_{glb}^{t-2}-\boldsymbol{w}_{n\_loc}^{t-2}\right\|+\eta g_{m a x}\left(\boldsymbol{w}_{glb}^{t-2}\right) \sum_{i=1}^C\left\|p^{(k)}(y=i)-p(y=i)\right\| \\
\leq & \left(a_{n}\right)^2\left\|\boldsymbol{w}_{glb}^{t-3}-\boldsymbol{w}_{n\_loc}^{t-3}\right\|+\eta \sum_{i=1}^C\left\|p^{(k)}(y=i)-p(y=i)\right\|\left(g_{\max }\left(\boldsymbol{w}_{glb}^{t-2}\right)+a_{n} g_{\max }\left(\boldsymbol{w}_{glb}^{t-2}\right)\right) \\
\leq & \left(a_{n}\right)^{t-1}\left\|\boldsymbol{w}_{glb}^{(m-1) t}-\boldsymbol{w}_{n\_loc}^{(m-1) t}\right\|+\eta \sum_{i=1}^C\left\|p^{(k)}(y=i)-p(y=i)\right\|\left(\sum_{j=0}^{t-2}\left(a_{n}\right)^j g_{\max }\left(\boldsymbol{w}^{t-2-j}_{glb}\right)\right) \\
= & \left(a_{n}\right)^{t-1}\left\|\boldsymbol{w}_{glb}^{(m-1) t}-\boldsymbol{w}_{n\_loc}^{(m-1) t}\right\|+\eta \sum_{i=1}^C\left\|p^{(k)}(y=i)-p(y=i)\right\|\left(\sum_{j=0}^{t-2}\left(a_{n}\right)^j g_{\max }\left(\boldsymbol{w}^{t-2-j}_{glb}\right)\right)
\end{aligned}
\end{equation*}
\end{multicols}

Therefore,
\begin{multicols}{2}
\begin{equation*}
    \centering
    \begin{aligned}
    &\left\|\boldsymbol{w}_{glb}^{t} -\boldsymbol{w}_{n\_loc}^{t}\right\|  \leq  q_n^t*[\sum_{n=1}^N a^{t}_{n}*\left\|\boldsymbol{w}^{t-1}_{glb}-\boldsymbol{w}^{t-1}_{n\_loc}\right\| \\
    & +\eta   \sum_{n=1}^N\sum_{m=1}^M\left\|P_{n}(m) - P(m)\right\| \sum_{j=1}^{t-1}a^{i}_{n}*g_{\max }\left(\boldsymbol{w}^{t-1-n}_{n\_loc}\right) ]
\end{aligned}
\end{equation*}
\end{multicols}
Hence proved.